\documentclass[letterpaper, 10 pt, conference]{ieeeconf}  

\IEEEoverridecommandlockouts                              

\usepackage{graphics} 
\usepackage{epsfig} 
\usepackage{mathptmx} 
\usepackage{times} 
\usepackage{amsmath} 
\usepackage{amssymb}  
\usepackage{pifont}
\usepackage{booktabs}
\usepackage{hyperref}
\usepackage{xcolor}
\usepackage{tablefootnote}
\usepackage{threeparttable}

\title{\LARGE \bf
Temporal Generalization in fNIRS-Based Autism Classification: \\ A Cross-Time-Window Transfer Benchmark
}

\author{Marios Petrov$^{1}$, Sahana Vinayak$^{1}$, Targol Bakhtiarvand$^{1}$,   Moses Smith Guddah$^{1}$, 
\\ Adham Atyabi$^{1}$, Frederick Shic$^{2,3}$, and Kevin A. Pelphrey$^{2}$%
\thanks{$^{1}$Department of Computer Science, University of Colorado Colorado Springs, Colorado Springs, CO 80918, USA}%
\thanks{$^{2}$Center for Child Health, Behavior and Development, Seattle Children's Research Institute, Seattle, WA, USA}%
\thanks{$^{3}$Department of Pediatrics, University of Washington School of Medicine, Seattle, WA, USA}%
}

\begin{document}

\maketitle
\thispagestyle{empty}
\pagestyle{empty}

\begin{abstract}
Functional near-infrared spectroscopy (fNIRS) is a promising modality for autism spectrum disorder (ASD) classification, yet existing approaches assume temporally aligned evaluation. In practice, the optimal observation window varies across subjects due to differences in hemodynamic delay and neurovascular coupling, creating a temporal distribution shift that degrades performance. We formalize this as a \textit{cross-time-window transfer problem}, introducing a protocol that varies window length (2.5--10\,s) and offset within biological motion trials. Using topographic map representations of fNIRS recordings, we benchmark three vision architectures under two zero-shot baselines and eight adaptation strategies under leave-one-subject-out cross-validation ($N{=}124$). Key findings: (1) zero-shot cross-window accuracy is near chance (54--69\%); (2) ${\approx}5\%$ subject-specific fine-tuning recovers 90--96\%, while a subject-specific upper bound reaches 97--100\%, identifying inter-subject variability as the dominant barrier; (3) domain-adversarial and self-supervised strategies achieve 78--90\% without target-subject data; and (4) discriminative information is recoverable from windows as short as 2.5\,s. These findings provide a practical roadmap for deploying fNIRS-based ASD classifiers under realistic temporal variability.
\end{abstract}
\section{INTRODUCTION}

Autism spectrum disorder (ASD) is a heterogeneous neurodevelopmental condition characterized by differences in social communication and restricted, repetitive behaviors~\cite{Hodges2020Autism}. Early identification is critical, as timely access to intervention can substantially improve developmental outcomes for affected children and their families~\cite{Okoye2023Early}. However, current diagnostic practice relies heavily on behavioral observation, clinical interviews, and parent-report instruments---processes that are time-intensive, subjective, and inequitably distributed, leaving many children in underserved communities undiagnosed or diagnosed late~\cite{Falkmer2013Diagnostic}. This has motivated sustained interest in objective, neurophysiologically grounded biomarkers that could support scalable ASD screening.

Functional near-infrared spectroscopy (fNIRS) is well suited to this goal. It is portable, tolerant of motion, safe for pediatric populations, and capable of monitoring cortical hemodynamics during naturalistic tasks~\cite{Wang2025The}. Over the past decade, fNIRS studies have consistently identified atypical hemodynamic signatures associated with ASD across prefrontal, temporal, and parietal regions during social and cognitive paradigms~\cite{Zhang2019Exploring, Liu2017Assessing, Conti2022Looking}. More recently, deep learning methods---including graph-based models, attention mechanisms, spatiotemporal convolution networks, and CNN--transformer hybrids---have achieved promising ASD-versus-TD classification accuracy from fNIRS signals~\cite{Cai2025Classification, Zhang2023Identification, Wang2022Transformer, Liao2024CTNet}. A recent systematic review of machine learning and deep learning applied to EEG and fNIRS for ASD diagnosis confirms the growing maturity of the field while highlighting persistent challenges in preprocessing heterogeneity and cross-site generalization~\cite{DeGiacomo2026Machine}. These advances suggest that fNIRS carries sufficient information for automated diagnostic support when paired with appropriate computational models.

Biological motion (BM) paradigms, which present brief point-light displays of human movement, are particularly attractive stimuli for fNIRS-based ASD classification. They reliably elicit differential neural responses between ASD and TD groups in prefrontal and temporal cortices~\cite{Mazziotti2022The, Scaffei2023A}. Recent work using high-density diffuse optical tomography---an advanced optical neuroimaging modality closely related to fNIRS---has confirmed that biological motion perception produces robust, regionally specific activation differences between autistic and non-autistic children in visual, motor, and social processing areas, with brain function correlating dimensionally with autism trait severity~\cite{Yang2024Mapping}. Complementary EEG evidence further demonstrates that biological motion elicits segregated, predominantly local network dynamics in ASD compared to the integrated long-range connectivity observed in typically developing controls~\cite{Bohm2024Segregated}. The short, repeatable trial structure of BM paradigms enables dense temporal sampling of the hemodynamic response within a single recording session, making BM-fNIRS a promising combination for practical screening.

Despite this progress, a critical and largely overlooked challenge limits the deployment of fNIRS-based classifiers: \textbf{temporal distribution shift}. The hemodynamic response function (HRF) varies across individuals in peak latency, shape, and amplitude due to differences in neurovascular coupling, cortical anatomy, and age. Recent multimodal EEG-fNIRS work has shown that subject-specific modeling of the HRF through parametric estimation significantly improves the characterization of neurovascular coupling, underscoring the extent of inter-individual variability even within healthy populations performing identical tasks~\cite{Lin2024Subject}. Most existing classification studies train and evaluate on a single, fixed observation window---implicitly assuming that the chosen temporal segment generalizes across subjects and sessions. When this assumption is violated, as it inevitably is in heterogeneous populations such as children with ASD, classifier performance can degrade substantially~\cite{Jing2023Transformer}. This problem is compounded by a broader perception that fNIRS lacks the temporal resolution needed for fine-grained neural decoding due to the inherent sluggishness of the hemodynamic response. Yet if discriminative information is present in short fNIRS segments---and if its temporal location within a trial varies meaningfully across individuals---then the limitation may lie not in the modality itself, but in evaluation protocols that fail to account for this variability.

In this work, we formalize this challenge as a \textbf{cross-time-window transfer problem} for fNIRS-based ASD classification. We convert multichannel fNIRS recordings from a biological motion paradigm into two-dimensional topographic maps and treat classification as an image recognition task, benchmarking three architectures that span convolutional and hybrid transformer designs: EfficientNet-b0, EfficientNet-b3~\cite{Tan2019EfficientNet}, and MaxViT~\cite{Tu2022MaxViT}. Our central contribution is a systematic evaluation protocol that varies both window length (2.5\,s--10\,s) and temporal offset within the trial, training on one time window and evaluating on another under leave-one-subject-out cross-validation. Within this framework, we evaluate two zero-shot baselines and define eight adaptation strategies---spanning few-shot subject personalization, cohort-level window adaptation, domain-adversarial invariance learning, and self-supervised temporal pretraining---providing, to our knowledge, the first unified benchmark of temporal generalization strategies for fNIRS-based classification.

Our principal findings are as follows:
\begin{enumerate}
    \item Cross-time-window transfer is substantially harder than same-window cross-subject generalization, confirming that temporal shift is a first-order problem in fNIRS classification.
    \item Even minimal subject-specific fine-tuning (${\approx}5\%$ of samples) largely closes the performance gap, demonstrating the value of lightweight personalization.
    \item Domain-adversarial and self-supervised pretraining strategies offer competitive generalization without requiring any target-subject data, providing viable alternatives when subject-specific calibration is impractical.
    \item Discriminative ASD-versus-TD information is recoverable from fNIRS windows as short as 2.5 seconds, challenging the assumption that hemodynamic signals are too temporally coarse for short-segment classification.
\end{enumerate}

Together, these results provide a practical roadmap for deploying fNIRS-based ASD classifiers under the realistic temporal variability encountered in clinical and field settings.

\section{RELATED WORK}

Several lines of prior work are closely aligned with the present study, spanning fNIRS-based ASD classification, biological motion as a neural probe, deep learning architectures for fNIRS, and cross-subject transfer strategies. We review each in turn and conclude by positioning our contributions relative to the existing literature.

\subsection{fNIRS as a Biomarker Modality for ASD}

Reviews published in the early years of clinical fNIRS in ASD consistently positioned the technique as a promising modality for studying the condition, citing evidence for atypical activation and connectivity in prefrontal and temporal \textit{social brain} regions and the technique's portability and suitability for infants and young children who must remain awake and socially engaged~\cite{Zhang2019Exploring, Liu2017Assessing, Wang2025The}. A more recent systematic review focused exclusively on preschool-aged children synthesized thirteen clinical fNIRS studies and described convergent evidence for disrupted resting-state connectivity and task-evoked responses, arguing explicitly for a fNIRS ``signature'' that could serve as a biomarker for early detection or treatment monitoring~\cite{Conti2022Looking}. A 2026 systematic review of machine learning and deep learning applied to EEG and fNIRS for ASD diagnosis further confirms the growing maturity of the field while highlighting persistent heterogeneity in preprocessing pipelines and classification models, and a lack of cross-site and cross-session generalization studies~\cite{DeGiacomo2026Machine}.

Closer to the present aims, several studies have targeted quantitative biomarkers and automated classification. Using simple visual stimuli, Mazziotti et al. showed that occipital hemodynamic response amplitude and variability correlate with autistic traits in typically developing children~\cite{Mazziotti2022The}, while Scaffei et al. demonstrated that hemodynamic response amplitude and lateralization can distinguish female preschoolers with ASD from matched peers~\cite{Scaffei2023A}. These works establish the feasibility of short, child-friendly visual protocols for extracting discriminative fNIRS features related to social-communication differences.

\subsection{Biological Motion Paradigms in ASD Neuroimaging}

Biological motion (BM) paradigms using point-light displays of human movement are among the most well-established probes of social perception in ASD. Recent optical neuroimaging work by Yang et al. used high-density diffuse optical tomography to image 46 ASD and 49 non-autistic school-age children during coherent and scrambled BM viewing, finding significantly stronger activation contrast in non-autistic participants across visual, motor, and social processing areas, with brain function correlating dimensionally with autism trait severity as measured by the SRS-2~\cite{Yang2024Mapping}. Complementary EEG work by B{\"o}hm et al. demonstrated that BM perception elicits segregated, predominantly local feedforward network dynamics in ASD, in contrast to the integrated long-range bidirectional connectivity observed in typically developing controls, with SVM classification successfully distinguishing the two groups~\cite{Bohm2024Segregated}. Together, these studies confirm that BM paradigms produce robust, multi-modal neural signatures of ASD that are well suited for classification.

\subsection{Deep Learning Architectures for fNIRS Classification}

Deep learning methods for fNIRS classification have evolved rapidly. For ASD-specific work, Zhang et al. trained an adaptive spatiotemporal graph convolution network on resting-state fNIRS from bilateral frontal and temporal cortex, achieving 95.4\% accuracy using time windows as short as 2.1 seconds and demonstrating that brief hemodynamic segments can encode rich pathological information~\cite{Zhang2023Identification}. Cai et al. subsequently achieved 97.9\% accuracy with an edge-weight-enhanced graph attention network fusing node and edge features from resting-state fNIRS over both temporal lobes, outperforming classical ML and CNN baselines~\cite{Cai2025Classification}. Both studies, however, use resting-state datasets and train and evaluate within the same time window, leaving open the question of temporal generalization.

Beyond ASD, several architectures have advanced fNIRS classification more broadly. Wang et al. introduced fNIRS-T and fNIRS-PreT, demonstrating that self-attention over lightly preprocessed fNIRS channels outperforms CNNs and LSTMs on multiple motor BCI datasets~\cite{Wang2022Transformer}. The same group later proposed fNIRSNet, which explicitly incorporates delayed hemodynamic response characteristics into convolutional kernel design, achieving strong subject-specific and subject-independent performance with only 498 parameters---highlighting the importance of encoding HRF domain knowledge into model architecture~\cite{Wang2023fNIRSNet}. Liao et al. presented CT-Net, a CNN-Transformer fusion network that provides interpretable fNIRS classification through attention visualization~\cite{Liao2024CTNet}. Guglielmini et al. used a transformer encoder to reconstruct short-separation (extracerebral) signals from long-separation channels, enabling hardware-independent denoising that generalizes across systems and tasks~\cite{Guglielmini2025Transformer-based}. More broadly, transformer-based foundation models trained on large-scale fMRI data have shown promising zero-shot and fine-tuned results for mental-state decoding and ASD diagnosis, suggesting that pretrained representations can transfer across neuroimaging contexts~\cite{Yang2025A}.

\subsection{Cross-Subject Transfer and Few-Shot Adaptation}

Cross-subject generalization remains a central challenge for fNIRS-based classification due to inter-individual variability in hemodynamic response characteristics, optode-scalp coupling, and cortical anatomy. Jing et al. applied a transformer architecture for cross-subject mental workload classification from fNIRS, reporting substantial performance degradation when models trained on one cohort were evaluated on new subjects without adaptation~\cite{Jing2023Transformer}. Lyu et al. proposed domain adaptation methods based on distribution alignment to improve cross-subject and cross-session workload classification from fNIRS, demonstrating that explicit handling of inter-subject distributional shift can recover much of the within-subject performance~\cite{Lyu2021Domain}. More recently, Feng et al. introduced a heterogeneous transfer learning model for cross-subject fNIRS classification in stroke rehabilitation, achieving 83--91\% accuracy using leave-one-subject-out cross-validation across eight patients by transferring knowledge from EEG source domains~\cite{Feng2025Heterogeneous}. Recent EEG-fNIRS work by Lin et al. has further shown that subject-specific parametric modeling of the hemodynamic response function significantly improves the characterization of neurovascular coupling, quantifying the extent of inter-individual HRF variability that any cross-subject model must contend with~\cite{Lin2024Subject}.

In the few-shot regime, Jung and An proposed EFRM, a multimodal EEG-fNIRS representation-learning model for few-shot brain-signal classification that leverages shared latent representations across modalities to classify with limited labeled samples~\cite{Jung2025EFRM:}. However, this approach requires concurrent EEG recording and does not address temporal distribution shift within fNIRS alone.

\subsection{Positioning of the Present Work}

The present work is most closely related to Zhang et al.~\cite{Zhang2023Identification} and Cai et al.~\cite{Cai2025Classification} in using short-window fNIRS for ASD discrimination, and to the fNIRS-T/fNIRS-PreT and fNIRSNet lines of work in adopting architectures that account for hemodynamic response properties~\cite{Wang2022Transformer, Wang2023fNIRSNet}. However, prior fNIRS-ASD classification studies uniformly train and evaluate within the same temporal window, and cross-subject transfer work in fNIRS has focused on subject distributional shift without considering temporal distributional shift. Our approach differs in three key respects. First, we use a naturalistic biological motion paradigm rather than resting state, producing task-evoked hemodynamic responses with clinically meaningful temporal structure. Second, we formalize and systematically evaluate cross-time-window transfer as a distinct generalization challenge, varying both window length and temporal offset under leave-one-subject-out cross-validation. Third, we benchmark nine adaptation strategies---spanning zero-shot, few-shot, cohort-level adaptation, domain-adversarial invariance, self-supervised pretraining, and meta-learning---within a single unified framework, providing the first systematic comparison of temporal generalization strategies for fNIRS-based classification.

\section{METHODS}

\begin{figure*}[t]
    \centering
    \includegraphics[width=\textwidth]{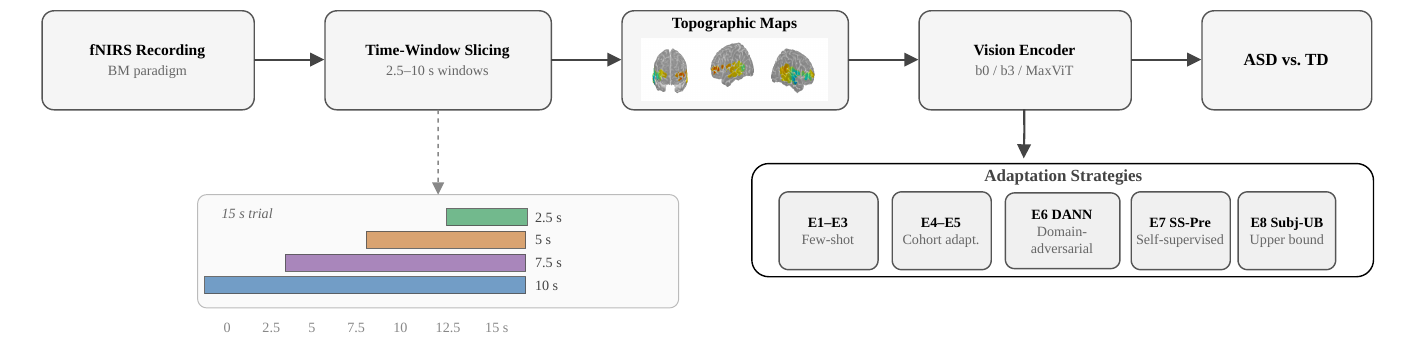}
    \caption{Overview of the cross-time-window transfer pipeline. Multichannel fNIRS recordings from a biological motion paradigm are segmented into overlapping time windows of varying length (2.5--10\,s) and offset. Each window is converted into a 2D topographic map via spatial interpolation, then classified by a vision encoder. Eight adaptation strategies (E1--E8) modify the training or fine-tuning procedure to address cross-subject and cross-window distribution shift.}
    \label{fig:pipeline} 
\end{figure*}

\subsection{Dataset}

\subsubsection{Participants}
We analyze data from $N{=}124$ children (ASD${=}63$, TD${=}61$), ages 7--12, with normal or corrected-to-normal vision and no neurological disorders beyond ASD. ASD diagnoses were confirmed via clinical assessment. Procedures were approved by the institutional IRB; guardians provided informed consent.

\subsubsection{Biological Motion Paradigm}
Participants viewed 15\,s point-light walker (PLW) clips across five conditions (happy, angry, fearful, neutral, and a rotation control). PLWs preserve biologically meaningful motion while removing facial features and background semantics. Onset and offset markers were aligned to the recording clock. Trials were excluded if $>$35\% of concurrent behavioral monitoring samples were invalid or if pre-trial quality checks indicated signal loss. Approximately 15\% of trials were removed; reported participant counts reflect the final post-filtering cohort.

\subsubsection{fNIRS Acquisition}
All fNIRS data were collected at Seattle Children's Hospital (SCH) as part of a broader neurodevelopmental research protocol. Device-specific acquisition parameters (system model, wavelengths, channel configuration, sampling rate, and optode layout) are documented in the original data-collection protocol and fall outside the scope of the present work, which operates on preprocessed channel-level hemodynamic time series.

\subsubsection{Preprocessing}
For each trial, the recorded fNIRS time series was segmented into various 10s, 7.5s, 5s, and 2.5s time windows. Within each window, the hemodynamic response was averaged across time points to produce a single scalar value per channel, yielding a spatially resolved snapshot of cortical activation over the entire montage. These channel-level averages were then mapped to scalp coordinates and interpolated into two-dimensional topographic images (see below). Artifact removal was done prior to image-generation.

\subsection{Problem Formulation and Topographic Map Generation}

Let $\mathcal{S} = \{S_1, S_2, \ldots, S_N\}$ denote the set of $N$ subjects, each labeled $y_i \in \{\text{ASD}, \text{TD}\}$. For a given trial duration $T$ (e.g., 15\,s), we define a \textit{time window} $w = [t_{\text{start}},\, t_{\text{start}} + \ell]$ parameterized by its onset $t_{\text{start}}$ and length $\ell \in \{2.5, 5, 7.5, 10\}\,$s. For each trial, the fNIRS time series is segmented according to $w$, and the mean hemodynamic concentration change is computed per channel. Channel-level values are mapped to their scalp coordinates and interpolated onto a regular two-dimensional grid, producing a topographic image $\mathbf{x} \in \mathbb{R}^{H \times W \times C}$. This converts the classification task into a standard image recognition problem. By varying $\ell$ and sliding $t_{\text{start}}$ in 2.5\,s increments, we generate a family of topographic maps per subject that tile the hemodynamic response at multiple resolutions and offsets.

We distinguish two temporal roles. The \textit{source window} $w_s$ generates the training data; the \textit{target window} $w_t \neq w_s$ is used to evaluate generalization under temporal distribution shift. For subject $S_i$ and window $w$, we write $\mathcal{D}_i^{w} = \{(\mathbf{x}_{i,j}^{w},\, y_i)\}_{j=1}^{n_i}$. The central question is: given a classifier $f_\theta$ trained on $\{\mathcal{D}_i^{w_s}\}$ from a subset of subjects, how well does it generalize to $\mathcal{D}_{S^*}^{w_t}$ from a held-out subject $S^*$?

\subsection{Architectures}

We benchmark three vision architectures: \textbf{EfficientNet-b0} and \textbf{EfficientNet-b3}~\cite{Tan2019EfficientNet}, compound-scaled convolutional networks serving as lightweight and higher-capacity baselines respectively; and \textbf{MaxViT}~\cite{Tu2022MaxViT}, a hybrid architecture combining multi-axis self-attention with MBConv layers to capture both local and global spatial dependencies. All models are initialized from ImageNet-pretrained weights and adapted with a binary classification head using AdamW with cosine annealing. Full hyperparameter details are provided in supplementary material.

\subsection{Leave-One-Subject-Out Evaluation}

All experiments follow leave-one-subject-out (LOSO) cross-validation. In each fold, one subject $S^*$ is held out; the base model is trained on source-window data from the remaining $N{-}1$ subjects:
\begin{equation}
    \theta^* = \arg\min_\theta \sum_{i \neq S^*} \mathcal{L}\bigl(f_\theta(\mathbf{x}),\, y_i\bigr), \quad \mathbf{x} \in \mathcal{D}_i^{w_s}
\end{equation}
where $\mathcal{L}$ is cross-entropy loss. This base model has never seen data from $S^*$ or from $w_t$.

\subsection{Zero-Shot Evaluation}

The base model $f_{\theta^*}$ is evaluated on $S^*$ under two conditions without further training. \textbf{Zero-shot source} ($Z_S$) evaluates on $\mathcal{D}_{S^*}^{w_s}$, measuring cross-subject generalization within the training window. \textbf{Zero-shot target} ($Z_T$) evaluates on $\mathcal{D}_{S^*}^{w_t}$, measuring simultaneous cross-subject and cross-window generalization. The gap $Z_S - Z_T$ quantifies the marginal cost of temporal distribution shift.

\subsection{Adaptation Strategies}

We define eight adaptation strategies (E1--E8) that progressively introduce subject-specific samples, cohort-level target-window data, invariance constraints, and self-supervised representations. In all few-shot conditions, ${\approx}5\%$ of the relevant sample pool is used. We denote the few-shot subset as $\mathcal{D}_{S^*}^{w,\text{fs}} \subset \mathcal{D}_{S^*}^{w}$, with $\mathcal{D}_{S^*}^{w,\text{eval}} = \mathcal{D}_{S^*}^{w} \setminus \mathcal{D}_{S^*}^{w,\text{fs}}$ reserved for evaluation.

\subsubsection{E1\texorpdfstring{\,}{}(FS-Src): Few-Shot Source-Window Personalization}
Fine-tune $f_{\theta^*}$ on $\mathcal{D}_{S^*}^{w_s,\text{fs}}$, then evaluate on $\mathcal{D}_{S^*}^{w_t}$. The model gains subject knowledge from $w_s$ but receives no exposure to $w_t$.

\subsubsection{E2\texorpdfstring{\,}{}(FS-Tgt): Few-Shot Target-Window Personalization}
Partition $S^*$'s target-window samples into $\mathcal{D}_{S^*}^{w_t,\text{fs}}$ (${\approx}5\%$) and $\mathcal{D}_{S^*}^{w_t,\text{eval}}$. Fine-tune $f_{\theta^*}$ on $\mathcal{D}_{S^*}^{w_t,\text{fs}}$ and evaluate on $\mathcal{D}_{S^*}^{w_t,\text{eval}}$. This provides an upper bound on what minimal labeled target-window data can achieve.

\subsubsection{E3\texorpdfstring{\,}{}(FS-Seq): Two-Stage Adaptation (Source $\rightarrow$ Target)}
Fine-tune $f_{\theta^*}$ on $\mathcal{D}_{S^*}^{w_s,\text{fs}}$ to obtain $f_{\theta_1}$, then fine-tune $f_{\theta_1}$ on $\mathcal{D}_{S^*}^{w_t,\text{fs}}$ with halved learning rate to obtain $f_{\theta_2}$. Evaluate on $\mathcal{D}_{S^*}^{w_t,\text{eval}}$. This tests whether a curriculum adapting first to the subject, then to the window, outperforms direct adaptation (E2).

\subsubsection{E4\texorpdfstring{\,}{}(Coh-Tgt): Cohort-Level Target-Window Adaptation}
Fine-tune the base model on all target-window samples from every subject except $S^*$:
\begin{equation}
    \theta_{\text{E4}} = \arg\min_\theta \sum_{i \neq S^*} \mathcal{L}\bigl(f_\theta(\mathbf{x}),\, y_i\bigr), \quad \mathbf{x} \in \mathcal{D}_i^{w_t}
\end{equation}
then evaluate on $\mathcal{D}_{S^*}^{w_t}$. This isolates the value of window adaptation without subject-specific exposure.

\subsubsection{E5\texorpdfstring{\,}{}(Coh+FS): Cohort + Few-Shot Personalization}
Extend E4 with a subject-specific stage: fine-tune $f_{\theta_{\text{E4}}}$ on $\mathcal{D}_{S^*}^{w_t,\text{fs}}$ and evaluate on $\mathcal{D}_{S^*}^{w_t,\text{eval}}$. This tests whether cohort-level window alignment and subject calibration are synergistic.

\subsubsection{E6\texorpdfstring{\,}{}(DANN): Domain-Adversarial Time-Window Invariance}
Train on data from all subjects except $S^*$ using both $w_s$ and $w_t$, with a classification loss $\mathcal{L}_{\text{cls}}$ and a domain discrimination loss $\mathcal{L}_{\text{dom}}$. A gradient-reversal layer~\cite{Ganin2016Domain} forces the feature extractor to suppress window-discriminative cues:
\begin{equation}
    \min_{\theta_f, \theta_c} \; \mathcal{L}_{\text{cls}} - \lambda \, \mathcal{L}_{\text{dom}}, \quad \min_{\theta_d} \; \mathcal{L}_{\text{dom}}
\end{equation}
where $\lambda$ controls adversarial strength. Evaluate on $\mathcal{D}_{S^*}^{w_t}$ in zero-shot mode.

\subsubsection{E7\texorpdfstring{\,}{}(SS-Pre): Self-Supervised Temporal Pretraining}
Pretrain the vision encoder on topographic maps from all subjects except $S^*$ across both windows using a temporal prediction objective (next-step embedding prediction). Fine-tune for ASD vs.\ TD classification on source-window data, then evaluate on $\mathcal{D}_{S^*}^{w_t}$. This tests whether self-supervised temporal representations improve cross-window robustness.

\subsubsection{E8\texorpdfstring{\,}{}(Subj-UB): Subject-Specific Upper Bound}
Fine-tune the base model on \textit{all} source-window samples from $S^*$:
\begin{equation}
    \theta_{\text{E8}} = \arg\min_\theta \mathcal{L}\bigl(f_\theta(\mathbf{x}),\, y_{S^*}\bigr), \quad \mathbf{x} \in \mathcal{D}_{S^*}^{w_s}
\end{equation}
then evaluate on $\mathcal{D}_{S^*}^{w_t}$. This upper bound tests within-subject cross-window transfer with full knowledge of $S^*$'s physiology but zero exposure to $w_t$.

\subsection{Summary of Experimental Conditions}

The eight strategies form a progression of increasing privileged information. The zero-shot baselines ($Z_S$, $Z_T$) use no data from the held-out subject $S^*$ or the target window. E1\,(FS-Src) receives a few-shot sample (${\approx}5\%$) from $S^*$'s source window only; E2\,(FS-Tgt) receives a few-shot sample from $S^*$'s target window only; E3\,(FS-Seq) receives both. E4\,(Coh-Tgt), E6\,(DANN), and E7\,(SS-Pre) use full target-window data from all other subjects but nothing from $S^*$; E5\,(Coh+FS) adds a few-shot target-window sample from $S^*$ on top of E4. E8\,(Subj-UB) uses all of $S^*$'s source-window data, providing the strongest subject personalization and serving as an upper bound.
\section{RESULTS}

We report accuracy for all eight adaptation strategies (E1--E8) and two zero-shot baselines ($Z_S$, $Z_T$) across three architectures under leave-one-subject-out cross-validation. Cells marked \textit{PH} denote pending experiments currently in progress for 2.5\,s window configurations.

\subsection{Zero-Shot Transfer}

Zero-shot performance is uniformly poor across all window configurations and architectures. Under same-window evaluation ($Z_S$), accuracy ranges from 54.0--65.1\% (mean: b0\,=\,57.0\%, b3\,=\,58.7\%, MaxViT\,=\,63.3\%). Under cross-window evaluation ($Z_T$), accuracy ranges from 55.4--69.5\% (mean: b0\,=\,63.9\%, b3\,=\,58.5\%, MaxViT\,=\,63.5\%). Neither window length nor offset produces a consistent advantage, and the $Z_S$--$Z_T$ gap is small and inconsistent in sign, indicating that temporal shift compounds rather than replaces the already-severe cross-subject degradation. These results confirm that cross-time-window transfer is a first-order challenge requiring explicit adaptation.

\subsection{Overview of Adaptation Strategies}

\begin{figure}[t]
    \centering
    \includegraphics[width=\columnwidth]{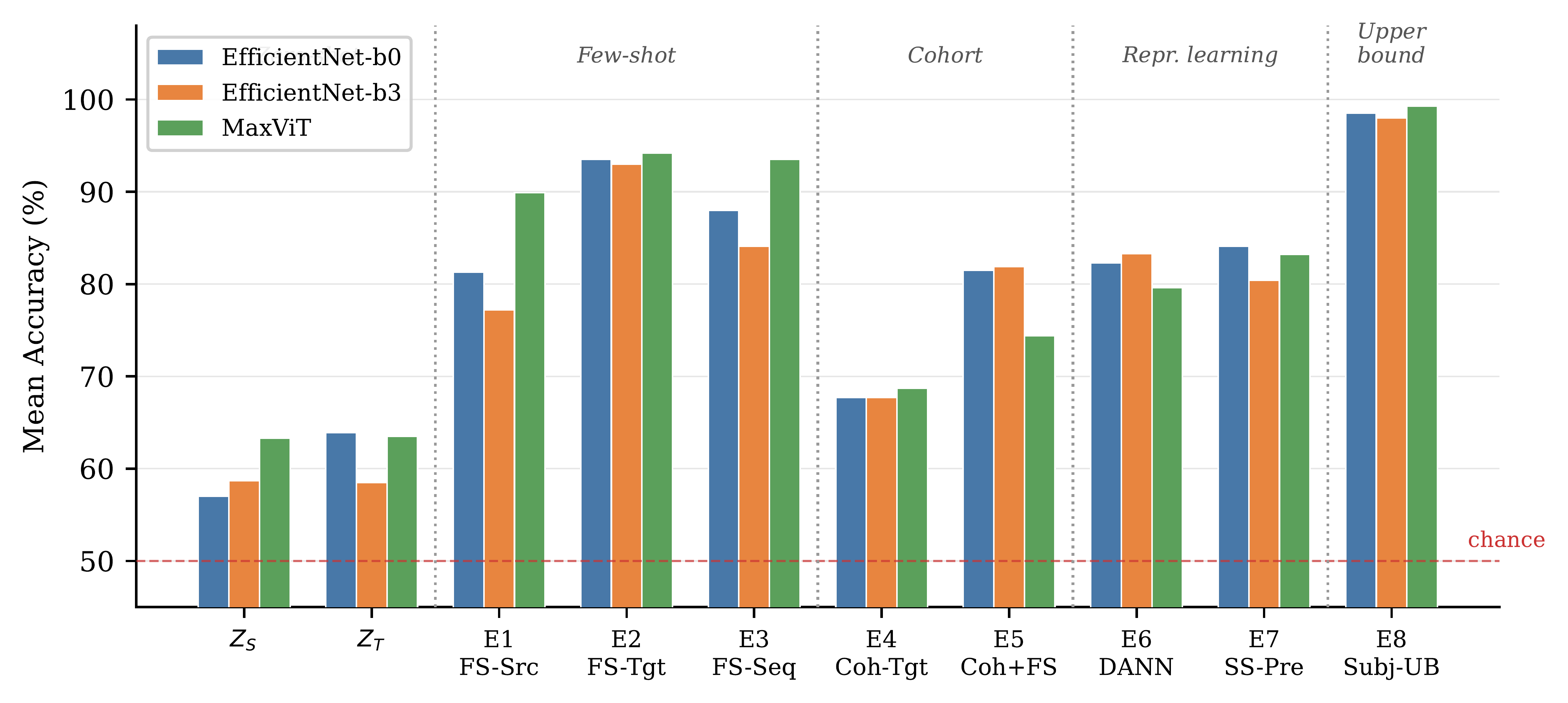}
    \caption{Mean accuracy (\%) across completed window configurations for each strategy and architecture. Zero-shot baselines cluster near chance (54--69\%). Few-shot target-window personalization (E2) recovers 90--96\%, while the subject-specific upper bound (E8) reaches 97--100\%. Among subject-independent strategies, E6\,(DANN) and E7\,(SS-Pre) at 78--90\% substantially outperform cohort-level adaptation (E4, ${\sim}$68\%). Best practical result across architectures is E2\,(FS-Tgt) with MaxViT at 94.2\%.}
    \label{fig:strategy_comparison}
\end{figure}

Figure~\ref{fig:strategy_comparison} summarizes mean accuracy per strategy averaged across all completed window configurations. E8\,(Subj-UB) is separated as an upper-bound reference.

Zero-shot baselines remain near chance (54--69\%), confirming temporal distribution shift as a first-order challenge. E2\,(FS-Tgt) achieves 93--94\% from only ${\approx}5\%$ target-window labels, substantially outperforming E1\,(FS-Src) (77--90\%). E8\,(Subj-UB) reaches 97--100\%, revealing that inter-subject variability, not the temporal shift, is the dominant barrier. Among subject-independent strategies, E6\,(DANN) and E7\,(SS-Pre) achieve 78--90\%, far exceeding E4\,(Coh-Tgt) (${\sim}$68\%).

\subsection{Comparison with Prior Work}

Table~\ref{tab:comparison} contextualizes our results against prior fNIRS classification methods. Existing ASD classification studies report high accuracy (95--98\%) but evaluate within the same time window and often without strict cross-subject holdout, making direct numerical comparison misleading. Our protocol is strictly harder: all results use leave-one-subject-out cross-validation with cross-time-window transfer on a substantially larger cohort ($N{=}124$). Under these conditions, our subject-independent strategies (E6, E7) achieve 83--84\% with zero target-subject data, and few-shot personalization (E2) recovers 94\%.

\begin{table}[htb]
\begin{scriptsize}
\begin{center}
\centering
\caption{Comparison with prior fNIRS classification methods. ``Cross-Subj'' indicates subject-holdout evaluation. ``Cross-Win'' indicates cross-time-window transfer. Best in \textbf{bold}; second best \underline{underlined}. $^\dagger$Uses ${\approx}5\%$ labeled target-window data from the held-out subject.}
\label{tab:comparison}
\setlength{\tabcolsep}{2pt}
\begin{tabular}{@{}lcccccc@{}}
\toprule
\textbf{Method} & \textbf{Task} & \textbf{$N$} & \textbf{Prot.} & \textbf{XS} & \textbf{XW} & \textbf{Acc\,(\%)} \\
\midrule
\multicolumn{7}{@{}l}{\textit{Prior work --- same-window evaluation}} \\
Zhang~\cite{Zhang2023Identification} (ASGCN)    & Rest & 47  & $k$-fold & \ding{55} & \ding{55} & 95.4 \\
Cai~\cite{Cai2025Classification} (EWE-GAT)      & Rest & 47  & LOOCV    & \ding{55} & \ding{55} & \underline{97.9} \\
\midrule
\multicolumn{7}{@{}l}{\textit{Prior work --- cross-subject transfer (non-ASD)}} \\
Feng~\cite{Feng2025Heterogeneous} (CHTLM)       & MI   & 8   & LOSO     & \ding{51} & \ding{55} & 83.1--91.3 \\
\midrule
\multicolumn{7}{@{}l}{\textit{Ours --- cross-subj.\ \textbf{and} cross-win.\ (LOSO, $N{=}124$)}} \\
E6\,(DANN, b3)              & BM & 124 & LOSO & \ding{51} & \ding{51} & 83.3 \\
E7\,(SS-Pre, b0)            & BM & 124 & LOSO & \ding{51} & \ding{51} & 84.1 \\
E2\,(FS-Tgt, MaxViT)$^\dagger$ & BM & 124 & LOSO & \ding{51} & \ding{51} & \underline{94.2} \\
\bottomrule
\end{tabular}
\end{center}
\end{scriptsize}
\end{table}

\subsection{Adaptation Strategies (E1--E8)}

Table~\ref{tab:all_strategies} reports per-window accuracy for all eight adaptation strategies.

\begin{table*}[t]
\centering
\caption{Accuracy (\%) for all adaptation strategies (E1--E8) across window configurations and architectures. E8\,(Subj-UB) is trained on all of $S^*$'s source-window data and serves as the subject-specific upper bound.}
\label{tab:all_strategies}
\resizebox{\textwidth}{!}{%
\begin{tabular}{cc|cccccccc|cccccccc|cccccccc}
\toprule
& & \multicolumn{8}{c|}{\textbf{EfficientNet-b0}} & \multicolumn{8}{c|}{\textbf{EfficientNet-b3}} & \multicolumn{8}{c}{\textbf{MaxViT}} \\
\textbf{Len} & \textbf{Win} & E1 & E2 & E3 & E4 & E5 & E6 & E7 & E8 & E1 & E2 & E3 & E4 & E5 & E6 & E7 & E8 & E1 & E2 & E3 & E4 & E5 & E6 & E7 & E8 \\
\midrule
10  & 5--15     & 82.4 & 94.8 & 90.5 & 67.4 & 80.8 & 84.2 & 83.5 & 98.4 & 79.3 & 94.2 & 84.9 & 67.8 & 82.6 & 81.4 & 78.9 & 98.1 & 92.0 & 95.8 & 95.1 & 68.7 & 74.1 & 80.6 & 82.3 & 99.5 \\
    & 2.5--12.5 & 82.0 & 95.1 & 90.8 & 67.2 & 79.3 & 83.6 & 81.7 & 98.7 & \underline{80.5} & 94.5 & 84.4 & 67.1 & 83.3 & 80.7 & 76.6 & 98.3 & \textcolor{blue}{\underline{93.7}} & 96.1 & \textcolor{blue}{\underline{96.2}} & 68.4 & 72.3 & 81.2 & 83.0 & 99.6 \\
    & 0--10     & 83.1 & 94.3 & \underline{91.2} & \underline{68.9} & 77.6 & 86.0 & 84.2 & 98.6 & 77.8 & 94.0 & 85.3 & 68.6 & 79.9 & 82.3 & 79.5 & 98.0 & 90.1 & 95.5 & 92.7 & 68.9 & 73.8 & 81.5 & 82.8 & 99.2 \\
\midrule
7.5 & 7.5--15   & 78.6 & 94.1 & 87.1 & 66.3 & 85.0 & 86.5 & 84.0 & 98.8 & 79.2 & 93.5 & 84.5 & 67.6 & 83.1 & 80.2 & 79.8 & 98.3 & 87.1 & 91.4 & 91.4 & 68.6 & 77.3 & 82.0 & 83.6 & 99.8 \\
    & 5--12.5   & 81.9 & 96.0 & 89.8 & 65.6 & 84.4 & 85.3 & 83.3 & 98.5 & 79.5 & 93.7 & 84.7 & 66.8 & 82.3 & 78.8 & 79.4 & 98.4 & 92.0 & 95.0 & 94.1 & 67.2 & 72.9 & 82.8 & 84.1 & 98.9 \\
    & 2.5--10   & 80.9 & 94.6 & 88.0 & 68.1 & 77.8 & 88.0 & 85.0 & 98.9 & 74.4 & \underline{95.0} & 82.5 & 66.5 & \underline{83.9} & 82.0 & 80.3 & 98.7 & 86.4 & 92.3 & 92.4 & 68.9 & 73.9 & 80.5 & 81.8 & 99.8 \\
    & 0--7.5    & 82.9 & \underline{96.3} & 89.6 & 68.9 & 78.7 & \textcolor{blue}{\underline{89.8}} & 85.8 & 97.9 & 79.1 & \underline{95.0} & 84.2 & 68.4 & 79.7 & 83.9 & \underline{84.2} & 97.8 & 91.9 & 95.8 & 95.0 & 68.9 & 72.1 & 83.4 & 79.5 & 99.0 \\
\midrule
5   & 10--15    & \underline{83.4} & 93.9 & 86.8 & 67.8 & 84.5 & 89.1 & 85.4 & \underline{99.3} & 77.5 & 93.8 & 84.2 & 67.9 & 82.1 & 86.4 & 82.4 & 98.1 & 89.5 & 93.2 & 92.4 & 68.6 & 76.0 & \underline{89.5} & \underline{88.5} & 99.4 \\
    & 7.5--12.5 & 82.6 & 96.1 & 88.2 & 68.8 & 82.8 & 78.1 & 81.8 & 98.7 & 78.8 & 94.2 & 84.8 & 67.3 & 81.6 & 84.0 & 78.1 & 98.7 & 90.0 & 92.5 & 93.8 & 68.9 & 76.3 & 76.6 & 83.4 & 99.6 \\
    & 5--10     & 80.9 & 95.4 & 86.9 & 66.2 & 82.2 & 80.8 & 85.4 & 97.6 & 79.3 & 93.9 & 82.8 & 67.4 & 81.0 & 82.9 & 79.9 & 96.9 & 88.0 & 90.3 & 89.5 & \underline{69.0} & 75.7 & 82.1 & 81.4 & 97.7 \\
    & 2.5--7.5  & 81.7 & 94.5 & 89.9 & 68.4 & 83.3 & 83.0 & 84.1 & 97.7 & 75.0 & 93.3 & 80.5 & 67.5 & 83.2 & \underline{87.6} & 80.3 & 96.7 & 89.7 & 95.4 & 91.6 & 68.9 & 73.1 & 78.3 & 78.6 & 99.6 \\
    & 0--5      & 81.8 & 95.4 & 88.1 & 67.9 & 80.2 & 78.6 & 86.4 & 99.0 & 76.8 & 93.6 & 87.0 & \textcolor{blue}{\underline{69.6}} & 82.7 & 80.9 & 81.7 & 98.8 & 90.4 & 94.0 & 91.9 & 68.9 & 78.3 & 79.0 & 83.1 & 99.6 \\
\midrule
2.5 & 12.5--15  & 78.2   & 94.2   & 89.6   & 68.6   & 80.3   & 79.8 & 85.2 & 98.8    & 77.0 & 90.6 & 83.7 & 67.3   & 79.5   & 86.8 & 82.6 & 97.8    & 93.3   & 96.1   & 94.8  & 68.4 & 74.7 & 77.3 & 84.7 & \textcolor{red}{\underline{100.0}}
 \\
    & 10--12.5  & 81.4 & 93.7 & 89.1 & 66.8   & \textcolor{blue}{\underline{86.3}}   & 79.0 & 83.9 & 97.6    & 78.3 & 92.5 & 86.8 & 67.1   & 80.2   & 85.0 & 78.8 & 98.3    & 91.4 & 94.8 & 94.3 & 67.8 & 78.1 & 75.1 & 84.0 & 98.1 \\
    & 7.5--10   & 78.7 & 87.7 & 84.5 & 68.8   & 80.7   & 80.3 & 81.0 & 97.3    & 74.6 & 90.3 & 82.5 & 67.2   & 80.3   & 79.6 & 81.1 & 97.2    & 90.9 & 94.6 & 94.6 & 68.5 & \underline{78.9} & 77.5   & 81.0   & 98.5 \\
    & 5--7.5    & 82.5 & 93.3 & 87.9 & 67.4   & 81.4   & 79.3 & 82.4 & 96.0    & 75.6 & 92.0 & \underline{87.2} & 67.2   & 81.1   & 83.0 & 78.5 & 98.1    & 86.8 & 93.1 & 93.5 & 68.9 & 76.0 & 77.4 & 79.3 & 98.7 \\
    & 2.5--5    & 79.5 & 90.1 & 87.9 & 67.9   & 81.3   & 81.0 & \textcolor{blue}{\underline{90.3}} & 96.4    & 73.2 & 91.3 & 81.4 & 67.0   & 79.1   & 87.1 & 81.3 & \underline{99.4}    & 88.7 & \textcolor{blue}{\underline{96.8}} & 94.4 & 68.9 & 76.3 & 78.0 & 85.6 & 99.1 \\
    & 0--2.5    & 82.3 & 90.7 & 88.3 & 67.7 & 79.9 & 80.9 & 81.6 & \underline{99.3}  & 76.7 & 92.9 & 85.6 & 68.1   & 82.2   & 81.0 & 81.0 & 98.0    & 92.6 & 95.0 & 94.1 & 67.2 & 76.3 & 74.8 & 83.1 & 98.6 \\
\bottomrule
\end{tabular}
}
\begin{tablenotes}
\item \begin{tiny}
$^a$ \textcolor{red}{Overall best performance achieved.} $^b$ \textcolor{blue}{Best performance per Experiment ($E_i$)} $^c$ \underline{Best performance per Experiment ($E_i$) \& per Classifier.}
\end{tiny}
\end{tablenotes}
\end{table*}

E2\,(FS-Tgt) leads consistently (90--96\%), confirming that 5\% of target-window labels is the most effective lightweight adaptation. E1\,(FS-Src) achieves 75--93\% from source-window data alone, showing partial cross-window transfer of subject physiology. E3\,(FS-Seq) falls between the two. MaxViT achieves the highest E1 accuracy, suggesting its attention mechanism captures more transferable subject representations. E4\,(Coh-Tgt) yields only marginal gains over zero-shot (${\sim}$67--69\%), confirming that window-level information without subject-level information is insufficient. E5\,(Coh+FS) recovers more (72--85\%) but remains below E2\,(FS-Tgt), suggesting the cohort-adapted initialization is a weaker starting point for personalization. E6\,(DANN) and E7\,(SS-Pre) achieve 78--90\% without any target-subject data, substantially improving over zero-shot. E6 tends to favor EfficientNet-b3 (reaching 87.6\%), while E7 is more uniform across architectures. Neither consistently dominates, suggesting explicit invariance enforcement and implicit temporal structure learning are complementary. E8\,(Subj-UB) achieves 97--100\% despite zero target-window exposure, confirming that the dominant barrier is inter-subject variability rather than temporal shift.
\section{DISCUSSION}

\subsection{Temporal Shift is Real but Recoverable}

The zero-shot results establish that cross-time-window transfer is a non-trivial challenge for fNIRS-based ASD classification. Models trained on one temporal segment and evaluated on another show accuracy in the 54--69\% range, only modestly above chance. This confirms that topographic maps from different time windows carry meaningfully different distributional statistics, reflecting variability in hemodynamic delay, signal-to-noise ratio, and the evolving shape of the HRF across the trial. This degradation occurs even when training and test subjects are drawn from the same population, isolating the temporal shift as a distinct source of performance loss.

The E8\,(Subj-UB) upper bound reveals that this shift is almost entirely recoverable when subject physiology is known: accuracy reaches 97--100\% across all architectures despite zero target-window exposure. The dominant barrier is therefore not the temporal shift itself but inter-subject variability. Once a model has internalized a subject's hemodynamic profile from one window, it transfers that knowledge to another with minimal loss. The practical implication is that calibration time should be invested in subject-specific data collection rather than in matching the exact temporal window used during training.

\subsection{The Value of Minimal Personalization}

E2\,(FS-Tgt) achieves 90--96\% accuracy from only ${\approx}5\%$ of the target subject's target-window samples, jumping from ${\sim}$60\% at zero-shot. This suggests that clinically useful accuracy is achievable with as little as a single short calibration block. E1\,(FS-Src), using the same fraction of data from the source window, reaches 75--93\%, demonstrating that subject-level information partially transfers across windows even without target-window exposure.

E3\,(FS-Seq) consistently falls between E1 and E2, suggesting that direct target-window adaptation is preferable to the sequential curriculum. The two-stage procedure likely overfits to the source-window distribution before encountering the target. E2's simplicity, a single fine-tuning step on a handful of labeled samples, makes it the most attractive option for clinical workflows that permit brief per-subject calibration.

\subsection{Cohort-Level Adaptation Alone is Insufficient}

E4\,(Coh-Tgt) yields only marginal improvement over zero-shot (${\sim}$67--69\%) despite access to the full target-window distribution from the rest of the cohort. This is consistent with the finding that subject variability, not window variability, is the primary bottleneck.

E5\,(Coh+FS) partially recovers (72--85\%) but underperforms E2\,(FS-Tgt), which starts from the source-window base model. This suggests that cohort-level target-window adaptation may degrade the model's starting point by overfitting to other subjects' characteristics, a form of negative transfer that few-shot personalization only partially corrects. The methodological takeaway is that naively pooling more target-distribution data does not substitute for subject-specific information.

\subsection{Domain-Adversarial and Self-Supervised Approaches}

E6\,(DANN) and E7\,(SS-Pre) achieve 78--90\% without any target-subject data, making them the strongest options when calibration is impractical. E6 removes window-specific cues via gradient reversal; E7 learns temporal structure through self-supervised prediction across consecutive windows. That neither consistently dominates suggests they capture complementary aspects of generalization. In context, Feng et al.~\cite{Feng2025Heterogeneous} reported 83--91\% for cross-subject classification on 8 stroke patients without temporal shift; our E6 and E7 achieve comparable accuracy on a larger pediatric cohort ($N{=}124$) under the additional burden of cross-window transfer.

\subsection{Architecture Effects}

MaxViT generally outperforms both EfficientNet variants in few-shot settings (E1--E3), consistent with its multi-axis self-attention capturing both local and global spatial dependencies. For E6\,(DANN), however, EfficientNet-b3 achieves the highest accuracy on several configurations, implying that architecture selection should be conditioned on the adaptation strategy. EfficientNet-b0 remains competitive throughout despite being the lightest model, which is encouraging for deployment on embedded or point-of-care devices.

\subsection{Short Windows Carry Discriminative Information}

Across all strategies, 2.5\,s windows achieve accuracy comparable to longer windows (5--10\,s) when paired with appropriate adaptation, aligning with Zhang et al.~\cite{Zhang2023Identification} who reported high accuracy from 2.1\,s segments under same-window evaluation. Shorter windows enable more trials per session, increase robustness to motion artifacts, and reduce recording time. For pediatric ASD screening, where session duration and child compliance are binding constraints, this substantially expands the feasible design space.

\subsection{Time Windowing as Ecologically Valid Data Augmentation}

Beyond evaluation, the sliding time-window approach offers a principled solution to data scarcity. Standard synthetic augmentation (SMOTE, noise injection, GANs) risks producing samples that violate physiological constraints. Time windowing provides an alternative grounded in the biophysics of the signal: by varying window length and offset, a single 15\,s trial yields up to six topographic maps differing in hemodynamic phase, amplitude, and noise. Every sample corresponds to a real segment of a real hemodynamic response with no interpolation required. The E8 results (97--100\%) confirm this diversity is learnable, and for small-sample neuroimaging studies this offers a scalable, assumption-free augmentation strategy.

\subsection{Limitations and Future Directions}

Some 2.5\,s configurations remain pending for E4--E8. The study uses a single paradigm at one site; multi-paradigm and multi-site generalization remains to be demonstrated. While our cohort ($N{=}124$, ages 7--12) is larger than prior fNIRS-ASD studies, extension to younger children is an important next step. The topographic map representation discards intra-window temporal dynamics; spatiotemporal architectures may capture richer information. Future work should combine E6 and E7 into a unified pipeline, extend the framework to multi-session settings, and integrate fNIRS with complementary modalities such as eye tracking, which was collected concurrently.
\section{CONCLUSION}

We formalized the cross-time-window transfer problem for fNIRS-based ASD classification and introduced a systematic evaluation protocol that varies window length and temporal offset under leave-one-subject-out cross-validation. Benchmarking three vision architectures on topographic maps from a biological motion paradigm, we draw four conclusions: (1) temporal distribution shift is a first-order challenge, with zero-shot cross-window accuracy near chance (54--69\%); (2) the shift is almost entirely recoverable through subject-specific adaptation --- E8\,(Subj-UB) reaches 97--100\%, and E2\,(FS-Tgt) recovers 90--96\% from only ${\approx}5\%$ of target-window labels --- establishing inter-subject variability as the dominant barrier; (3) when subject-specific data is unavailable, domain-adversarial invariance (E6\,(DANN), 78--90\%) and self-supervised pretraining (E7\,(SS-Pre), 78--90\%) substantially outperform cohort-level adaptation (${\sim}$68\%); and (4) discriminative information is recoverable from windows as short as 2.5\,s. Beyond evaluation, the sliding time-window protocol itself serves as an ecologically valid data augmentation strategy, generating physiologically grounded training samples without synthetic interpolation. Together, our adaptation hierarchy and augmentation framework provide a practical roadmap extensible to other fNIRS classification tasks and clinical populations facing analogous temporal generalization challenges.
\section*{ACKNOWLEDGMENT}

This study was funded by NIMH award K01MH104739.
\bibliographystyle{IEEEtran}
\bibliography{sec/references}

@article{Hodges2020Autism,title={Autism spectrum disorder: definition, epidemiology, causes, and clinical evaluation},author={Holly Hodges and Casey Fealko and Neelkamal Soares},journal={Translational Pediatrics},year={2020},volume={9},pages={S55 - S65},doi={10.21037/tp.2019.09.09}}

@article{Okoye2023Early,title={Early Diagnosis of Autism Spectrum Disorder: A Review and Analysis of the Risks and Benefits},author={Chiugo Okoye and Chidi M Obialo-Ibeawuchi and Omobolanle A Obajeun and S. Sarwar and C. Tawfik and M. Waleed and Asad Ullah Wasim and Iman Mohamoud and Adebola Y Afolayan and Rheiner N Mbaezue},journal={Cureus},year={2023},volume={15},doi={10.7759/cureus.43226}}

@article{Falkmer2013Diagnostic,title={Diagnostic procedures in autism spectrum disorders: a systematic literature review},author={T. Falkmer and Katie Anderson and Marita Falkmer and C. Horlin},journal={European Child \& Adolescent Psychiatry},year={2013},volume={22},pages={329-340},doi={10.1007/s00787-013-0375-0}}

@article{Wang2025The,title={The functional near infrared spectroscopy applications in children with developmental diseases: a review},author={Jing Wang and Zhuo Zou and Haoyu Huang and Jinting Wu and Xianzhao Wei and Shuyue Yin and Yingjuan Chen and Yun Liu},journal={Frontiers in Neurology},year={2025},volume={16},doi={10.3389/fneur.2025.1495138}}

@article{Cai2025Classification,title={Classification of Autism Spectrum Disorder Using Edge-Weight Enhanced Graph Attention Network With Multiple Features of Resting-State fNIRS Signals},author={Jingwen Cai and Xi Zeng and Jun Li},journal={Journal of Biophotonics},year={2025},volume={18},doi={10.1002/jbio.202500138}}

@inproceedings{Jing2023Transformer,title={Transformer Based Cross-Subject Mental Workload Classification Using FNIRS for Real-World Application},author={Yitao Jing and Weiqun Wang and Jiaxing Wang and Yuze Jiao and Kexin Xiang and Tianyu Lin and Weiguo Shi and Zengguang Hou},booktitle={2023 45th Annual International Conference of the IEEE Engineering in Medicine \& Biology Society (EMBC)},year={2023},pages={1-5},doi={10.1109/embc40787.2023.10341167}}

@article{Jung2025EFRM:,title={EFRM: A Multimodal EEG-fNIRS Representation-learning Model for few-shot brain-signal classification},author={Euijin Jung and Jinung An},journal={Computers in biology and medicine},year={2025},volume={199},pages={111292},doi={10.1016/j.compbiomed.2025.111292}}

@article{Zhang2019Exploring,title={Exploring brain functions in autism spectrum disorder: A systematic review on functional near-infrared spectroscopy (fNIRS) studies},author={Fen Zhang and H. Roeyers},journal={International Journal of Psychophysiology},year={2019},volume={137},pages={41-53},doi={10.1016/j.ijpsycho.2019.01.003}}

@article{Liu2017Assessing,title={Assessing autism at its social and developmental roots: A review of Autism Spectrum Disorder studies using functional near-infrared spectroscopy},author={Tao Liu and Xingchen Liu and Li Yi and Chaozhe Zhu and Patrick S. Markey and Matthew Pelowski},journal={NeuroImage},year={2017},volume={185},pages={955-967},doi={10.1016/j.neuroimage.2017.09.044}}

@article{Conti2022Looking,title={Looking for ``fNIRS Signature'' in Autism Spectrum: A Systematic Review Starting From Preschoolers},author={Eugenia Conti and Elena Scaffei and Chiara Bosetti and Viviana Marchi and Valeria Costanzo and V. Dell'Oste and Raffaele Mazziotti and L. Dell'Osso and C. Carmassi and F. Muratori and L. Baroncelli and S. Calderoni and R. Battini},journal={Frontiers in Neuroscience},year={2022},volume={16},doi={10.3389/fnins.2022.785993}}

@article{Mazziotti2022The,title={The amplitude of fNIRS hemodynamic response in the visual cortex unmasks autistic traits in typically developing children},author={Raffaele Mazziotti and Elena Scaffei and Eugenia Conti and Viviana Marchi and Riccardo Rizzi and G. Cioni and R. Battini and L. Baroncelli},journal={Translational Psychiatry},year={2022},volume={12},doi={10.1038/s41398-022-01820-5}}

@article{Scaffei2023A,title={A Potential Biomarker of Brain Activity in Autism Spectrum Disorders: A Pilot fNIRS Study in Female Preschoolers},author={Elena Scaffei and Raffaele Mazziotti and Eugenia Conti and Valeria Costanzo and S. Calderoni and Andrea Stoccoro and C. Carmassi and R. Tancredi and L. Baroncelli and R. Battini},journal={Brain Sciences},year={2023},volume={13},doi={10.3390/brainsci13060951}}

@article{Zhang2023Identification,title={Identification of autism spectrum disorder based on functional near-infrared spectroscopy using adaptive spatiotemporal graph convolution network},author={Haoran Zhang and Lingyu Xu and Jie Yu and Jun Li and Jinhong Wang},journal={Frontiers in Neuroscience},year={2023},volume={17},doi={10.3389/fnins.2023.1132231}}

@article{Wang2022Transformer,title={Transformer Model for Functional Near-Infrared Spectroscopy Classification},author={Z. Wang and Jun Zhang and Xiaochu Zhang and Peng Chen and Bing Wang},journal={IEEE Journal of Biomedical and Health Informatics},year={2022},volume={26},pages={2559-2569},doi={10.1109/jbhi.2022.3140531}}

@article{Guglielmini2025Transformer-based,title={Transformer-based deep learning model for predicting fNIRS short-channel signals},author={S. Guglielmini and Vittoria Banchieri and F. Scholkmann and Martin Wolf},journal={Neurophotonics},year={2025},volume={12},doi={10.1117/1.nph.12.4.045008}}

@article{Yang2025A,title={A Foundational fMRI Model for Representing Continuous Brain States},author={Li Yang and Lei Guo and Yixuan Yuan and Junwei Han and Xintao Hu and Tuo Zhang},journal={IEEE Journal of Biomedical and Health Informatics},year={2025},volume={29},pages={8361-8373},doi={10.1109/jbhi.2025.3569627}}

@inproceedings{Tan2019EfficientNet,title={EfficientNet: Rethinking Model Scaling for Convolutional Neural Networks},author={Mingxing Tan and Quoc V. Le},booktitle={Proceedings of the 36th International Conference on Machine Learning (ICML)},year={2019},pages={6105--6114}}

@inproceedings{Tu2022MaxViT,title={MaxViT: Multi-Axis Vision Transformer},author={Zhengzhong Tu and Hossein Talebi and Han Zhang and Feng Yang and Jiahui Yu and Yinxiao Li and Peyman Milanfar},booktitle={European Conference on Computer Vision (ECCV)},year={2022},pages={459--479},doi={10.1007/978-3-031-20053-3\_27}}

@article{Yang2024Mapping,title={Mapping neural correlates of biological motion perception in autistic children using high-density diffuse optical tomography},author={Dalin Yang and Alexandra M. Svoboda and Tessa G. George and Patricia K. Mansfield and Muriah D. Wheelock and Mariel L. Schroeder and Sean M. Rafferty and Arefeh Sherafati and Kalyan Tripathy and Tracy Burns-Yocum and Elizabeth Forsen and John R. Pruett and Natasha M. Marrus and Joseph P. Culver and John N. Constantino and Adam T. Eggebrecht},journal={Molecular Autism},year={2024},volume={15},number={1},pages={35},doi={10.1186/s13229-024-00614-4}}

@article{Lin2024Subject,title={Subject-Specific Modeling of {EEG-fNIRS} Neurovascular Coupling by Task-Related Tensor Decomposition},author={Jianeng Lin and Jiewei Lu and Zhilin Shu and Jianda Han and Ningbo Yu},journal={IEEE Transactions on Neural Systems and Rehabilitation Engineering},year={2024},volume={32},pages={452-461},doi={10.1109/TNSRE.2024.3355121}}

@article{Bohm2024Segregated,title={Segregated Dynamical Networks for Biological Motion Perception in the Mu and Beta Range Underlie Social Deficits in Autism},author={Petra B{\"o}hm and Nico Adelhoefer and Irina Jarvers and Jens Blechert and Tobias Renner and Christian Scharfenort},journal={Diagnostics},year={2024},volume={14},number={4},pages={408},doi={10.3390/diagnostics14040408}}

@article{DeGiacomo2026Machine,title={Machine learning and deep learning applied to {EEG} and {fNIRS} for early autism spectrum disorder diagnosis: a systematic review},author={Andrea {De Giacomo} and Fernanda Craig and Silvia Medicamento and Francesca Gradia and Davide Sardella and Alessia Costabile and Emilia Matera and Marco Turi},journal={Frontiers in Psychiatry},year={2026},volume={17},doi={10.3389/fpsyt.2026.1668914}}

@article{Liao2024CTNet,title={{CT-Net}: An interpretable {CNN-Transformer} fusion network for {fNIRS} classification},author={Lingxiang Liao and Jingqing Lu and Lutao Wang and Yongqing Zhang and Dongrui Gao and Manqing Wang},journal={Medical and Biological Engineering and Computing},year={2024},volume={62},number={10},pages={3233--3247},doi={10.1007/s11517-024-03138-4}}

@article{Wang2023fNIRSNet,title={Rethinking Delayed Hemodynamic Responses for {fNIRS} Classification},author={Zenghui Wang and Jihong Fang and Jun Zhang},journal={IEEE Transactions on Neural Systems and Rehabilitation Engineering},year={2023},volume={31},pages={4528--4538},doi={10.1109/TNSRE.2023.3330911}}

@article{Lyu2021Domain,title={Domain adaptation for robust workload level alignment between sessions and subjects using {fNIRS}},author={Boyang Lyu and Thao Pham and Giles Blaney and Zachary Haga and Angelo Sassaroli and Sergio Fantini and Shuchin Aeron},journal={Journal of Biomedical Optics},year={2021},volume={26},number={2},pages={022908},doi={10.1117/1.JBO.26.2.022908}}

@article{Feng2025Heterogeneous,title={Heterogeneous transfer learning model for improving the classification performance of {fNIRS} signals in motor imagery among cross-subject stroke patients},author={Jin Feng and YunDe Li and ZiJun Huang and Yehang Chen and SenLiang Lu and RongLiang Hu and QingHui Hu and YuYao Chen and XiMiao Wang and Yong Fan and Jing He},journal={Frontiers in Human Neuroscience},year={2025},volume={19},doi={10.3389/fnhum.2025.1555690}}

@article{Ganin2016Domain,title={Domain-Adversarial Training of Neural Networks},author={Yaroslav Ganin and Evgeniya Ustinova and Hana Ajakan and Pascal Germain and Hugo Larochelle and Fran{\c{c}}ois Laviolette and Mario Marchand and Victor Lempitsky},journal={Journal of Machine Learning Research},year={2016},volume={17},number={59},pages={1--35}}
\end{document}